\documentclass[journal]{IAENGtran}
\usepackage[pdftex]{graphicx}
\DeclareGraphicsExtensions{.pdf,.png,.jpg}

\usepackage[T1]{fontenc}
\usepackage[utf8]{inputenc}
\usepackage{lmodern}

\usepackage{newunicodechar}
\newunicodechar{−}{\ensuremath{-}}   
\newunicodechar{≤}{\ensuremath{\leq}} 
\newunicodechar{≥}{\ensuremath{\geq}} 

\usepackage{placeins}     
\usepackage{dblfloatfix}  

\usepackage{caption}
\usepackage[none]{hyphenat}   
\usepackage[hidelinks]{hyperref}
\begin{document}
%
\title{Quantifying Label-Induced Bias in Large Language Model Self- and Cross-Evaluations}
%
%
%

\author{Muskan~Saraf, Sajjad~Rezvani~Boroujeni, Justin~Beaudry, Hossein~Abedi, and Tom~Bush*%

\thanks{Muskan Saraf is with the Data Science Department, Actual Reality Technologies, Toledo, USA (e-mail: muskan@actualreality.tech).}%
\thanks{Sajjad Rezvani Boroujeni is with the Data Science Department, Actual Reality Technologies, Toledo, USA (e-mail: sajjadr@bgsu.edu).}%
\thanks{Justin Beaudry is with the Software Department, Actual Reality Technologies, Toledo, USA (e-mail: justin@actualreality.tech).}%
\thanks{Hossein Abedi is with the Data Science Department, Actual Reality Technologies, Toledo, USA (e-mail: hossein@actualreality.tech).}%
\thanks{Tom Bush is the CEO of Actual Reality Technologies, Maumee, USA (e-mail: tom@actualreality.tech). *Corresponding author.}}

\maketitle

\pagestyle{empty}
\thispagestyle{empty}

\begin{abstract}
Large language models (LLMs) are increasingly deployed as evaluators of text quality, yet the validity of their judgments remains underexplored. This study investigates systematic bias in self- and cross-model evaluations across three prominent LLMs: ChatGPT, Gemini, and Claude. We designed a controlled experiment in which blog posts authored by each model were evaluated by all three models under four labeling conditions: no attribution, true attribution, and two false-attribution scenarios. Evaluations employed both holistic preference voting and granular quality ratings across three dimensions Coherence, Informativeness, and Conciseness with all scores normalized to percentages for direct comparison. Our findings reveal pronounced asymmetries in model judgments: the "Claude" label consistently elevated scores regardless of actual authorship, while the "Gemini" label systematically depressed them. False attribution frequently reversed preference rankings, producing shifts of up to 50 percentage points in voting outcomes and up to 12 percentage points in quality ratings. Notably, Gemini exhibited severe self-deprecation under true labels, while Claude demonstrated intensified self-preference. These results demonstrate that perceived model identity can substantially distort both high-level judgments and fine-grained quality assessments, independent of content quality. Our findings challenge the reliability of LLM-as-judge paradigms and underscore the critical need for blind evaluation protocols and diverse multi-model validation frameworks to ensure fairness and validity in automated text evaluation and LLM benchmarking.
\end{abstract}

\begin{IAENGkeywords}
 Artificial Intelligence, Large language models, AI evaluation bias, label effects, cross-model evaluation, Benchmarking Fairness.
\end{IAENGkeywords}

%
\IAENGpeerreviewmaketitle

\section{Introduction}
%
%
%
%
\IAENGPARstart{L}{arge}  language models (LLMs) such as ChatGPT, Gemini, and Claude are increasingly deployed not only for content generation but also for content evaluation. This dual role raises a critical question: \textit{can LLMs evaluate outputs impartially, or are their judgments influenced by perceived authorship?} Previous studies have shown that both humans and models exhibit systematic bias, often favoring certain sources or stylistic patterns [1], [2]. When evaluators are aware of the source, their ratings may be shaped by prior expectations a phenomenon known as source bias [3]. In LLMs, this bias may manifest as self-preference bias, where a model rates its own outputs higher, or label-induced bias, where a model's name affects evaluation regardless of quality [4].

This study investigates these biases by analyzing how three leading LLMs ChatGPT-4o, Gemini 2.5 Flash, and Claude Sonnet 4 evaluate blog posts authored by themselves and each other under four controlled conditions: no labels, true labels, and two false-label scenarios. We employ two complementary scoring approaches: percentage-based preference scoring and point-based quality scoring for Coherence, Informativeness, and Conciseness, converted to percentages for direct comparison.

Our findings reveal striking asymmetries. The "Claude" label consistently boosts scores regardless of content, while the "Gemini" label consistently depresses them. False labels produce swings of up to 50 percentage points in preference scores and up to 12 percentage points in quality ratings. This work provides: (i) a controlled, multi-condition analysis of self- and cross-model evaluation bias, (ii) quantitative evidence comparing label effects across preference and quality dimensions, and (iii) recommendations for mitigating bias through blind or multi-model evaluation protocols.

The paper is organized as follows. Section II reviews related work. Section III describes our methodology. Section IV presents results. Section V discusses implications for LLM benchmarking. Section VI concludes with recommendations for future research.
 
\section{Related Work}
Bias in automated language model evaluation has garnered growing attention in recent years. Research on self-preference bias shows that LLMs favor their own outputs, with models demonstrating measurable self-recognition capabilities that correlate with stronger self-favoritism [5], [6].

Complementary work on label-induced evaluation bias reveals how LLMs may be swayed by perceived authorship regardless of content quality. Wang et al. demonstrate that systematic bias based on response position can manipulate rankings, even making weaker models outperform stronger ones under certain prompt orderings [7]. Chen et al. further investigate whether self-preference reflects genuine superiority or signaling bias, finding that harmful bias persists even in stronger models [8].

Researchers have also examined implicit versus explicit evaluation dynamics, revealing inconsistencies in how models consciously versus unconsciously express bias [9]. Similar concerns emerge across deep learning domains, where pre-trained and architecture-specific models despite high accuracy often inherit systematic biases from training regimes and structural choices [10], [11], [13], [14], [15]. These examples underscore a broader challenge: deep learning systems across language and vision domains are susceptible to implicit and structural biases that influence evaluation.

Broader surveys categorize bias into intrinsic and extrinsic types and emphasize mitigation strategies across data, model, and output layers [12]. These reviews underscore the relevance of our dual-method approach, which examines both overall preference and fine-grained quality criteria across controlled labeling experiments.

\section{Methodology}

This study investigates label-induced and self-preference bias in LLM evaluations using a controlled, multi-model, multi-condition design involving three stages: blog generation, evaluation under manipulated label conditions, and dual-method scoring analysis.

Three LLMs ChatGPT-4o, Gemini 2.5 Flash, and Claude Sonnet 4 generated blog posts using a fixed prompt template: \textit{"You are a professional blog writer. Write a concise blog post (around 200 words) for the title '<insert your title here>'. The style should be engaging and suitable for an online audience. Return only the blog content, no extra text."} Ten distinct titles covering diverse topics with similar complexity were used, with each model generating one blog per title, yielding 30 blog posts total.

Each model then evaluated all blogs including its own under four labeling conditions: no labels (no author attribution), true labels (correct attribution), False Label Scenario 1 (ChatGPT labeled as Gemini, Gemini as Claude, Claude as ChatGPT), and False Label Scenario 2 (ChatGPT labeled as Claude, Gemini as ChatGPT, Claude as Gemini).

Two scoring systems were applied. Percentage-based preference scores measured how often each output was chosen as "best" for a given title. Point-based quality scores rated Coherence, Informativeness, and Conciseness on a 0–10 scale, converted to percentages for direct comparability. Analyses were conducted at three levels: intra-condition (within-condition comparisons), cross-condition (tracking changes across conditions), and metric-specific (examining bias effects on each criterion). This design enabled controlled examination of both self-preference and label-induced bias in LLM evaluations.

\FloatBarrier

\section{\textbf{Results}}\label{AA}

\begin{figure}[htbp]
\centering
{\includegraphics[width=1\linewidth]{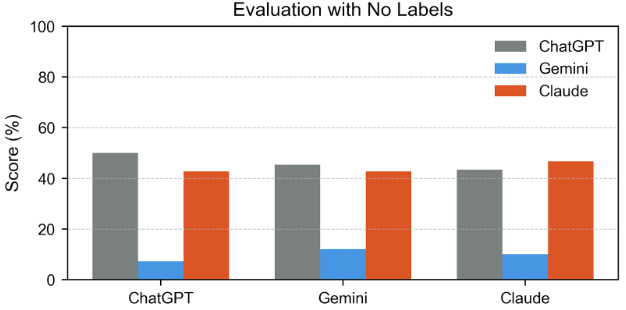}}
\caption{Percentage-based overall scores under the \textit{No Label} condition.}
\label{fig}
\end{figure}

\begin{figure}[htbp]
\centerline{\includegraphics[width=1\linewidth]{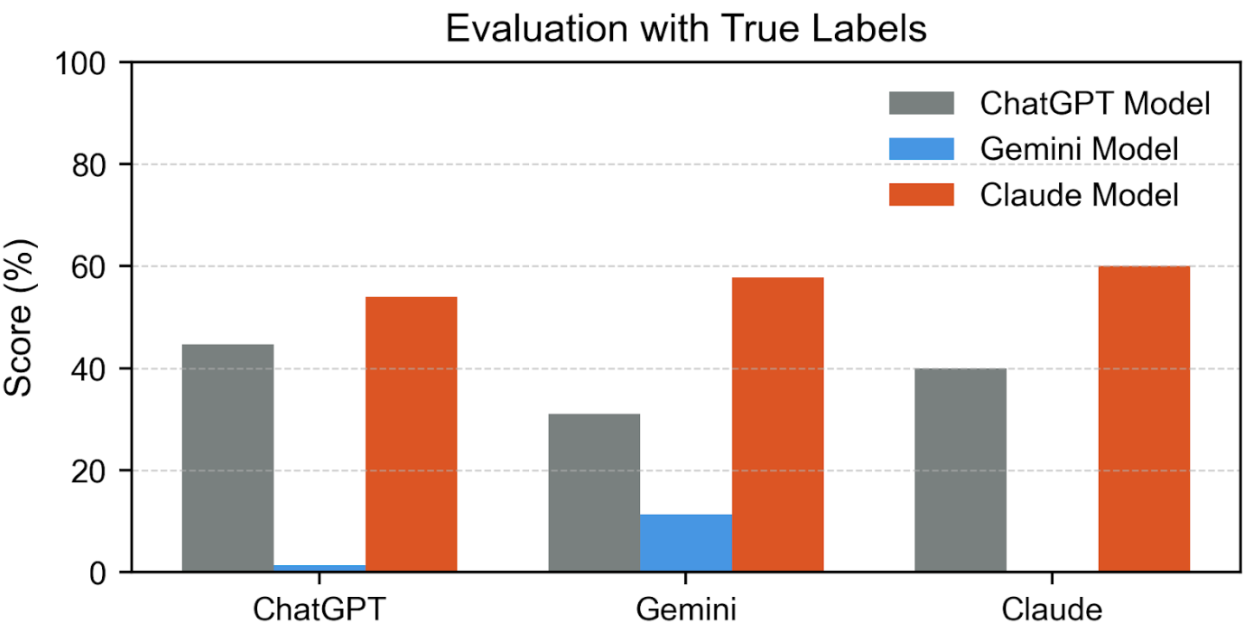}}
\caption{Percentage-based overall scores under the \textit{True Label} condition.}
\label{fig}
\end{figure}

\begin{figure}[htbp]
\centerline{\includegraphics[width=1\linewidth]{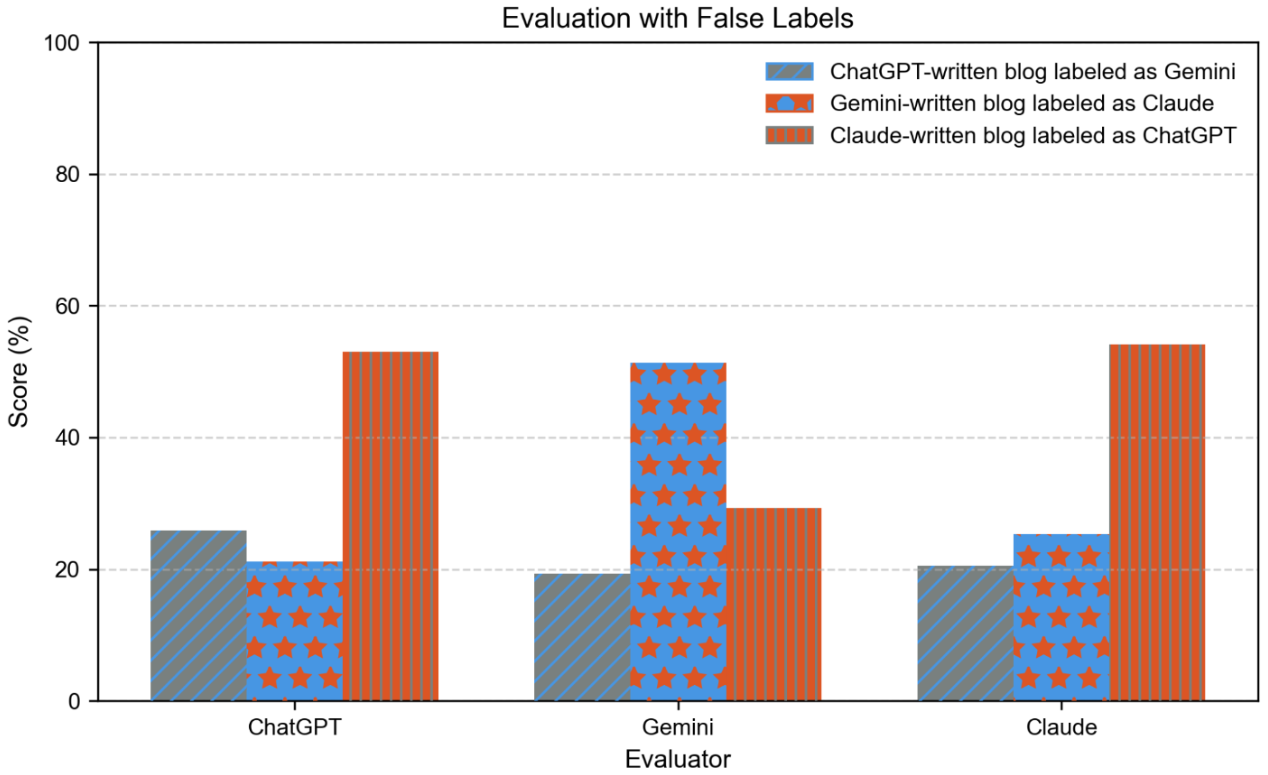}}
\caption{Percentage-based overall scores under \textit{False Label Scenario 1}
(ChatGPT-as-Gemini, Gemini-as-Claude, Claude-as-ChatGPT).}
\label{fig}
\end{figure}

\begin{figure*}[htbp]
\centering{\includegraphics[width=1\linewidth]{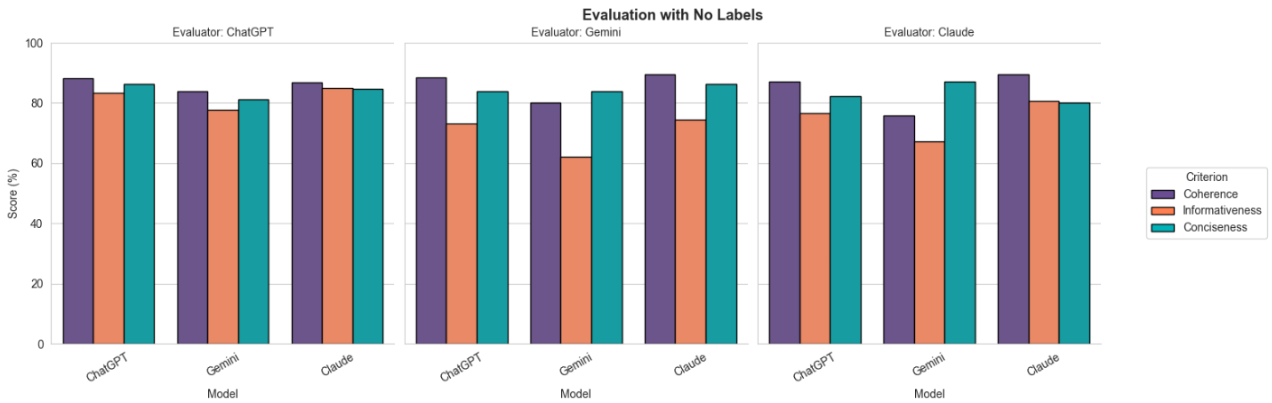}}
\caption{Point-based scores under the \textit{No Label} condition.}
\label{fig}
\end{figure*}

\begin{figure*}[htbp]
\centering{\includegraphics[width=1\linewidth]{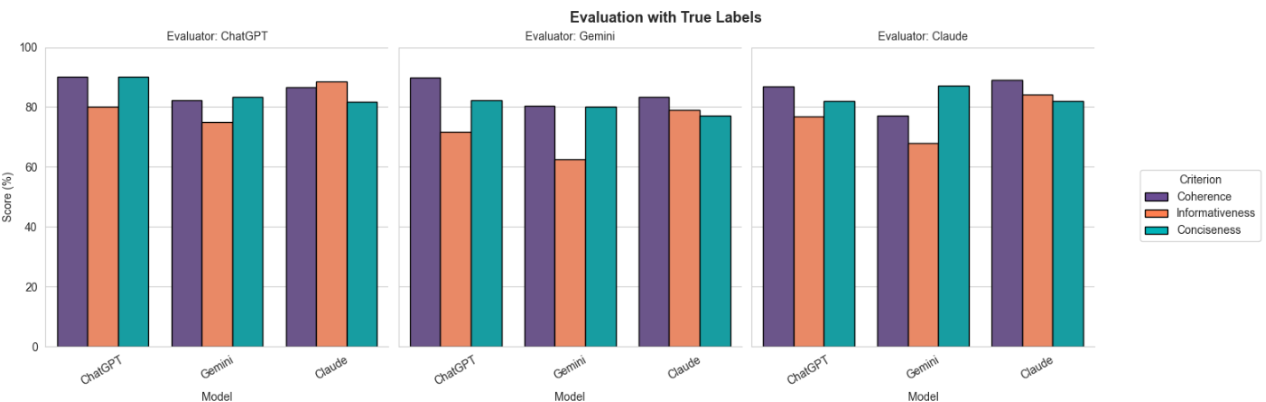}}
\caption{Point-based scores under the \textit{True Label} condition.}
\label{fig}
\end{figure*}

\begin{figure}[htbp]
\centerline{\includegraphics[width=1\linewidth]{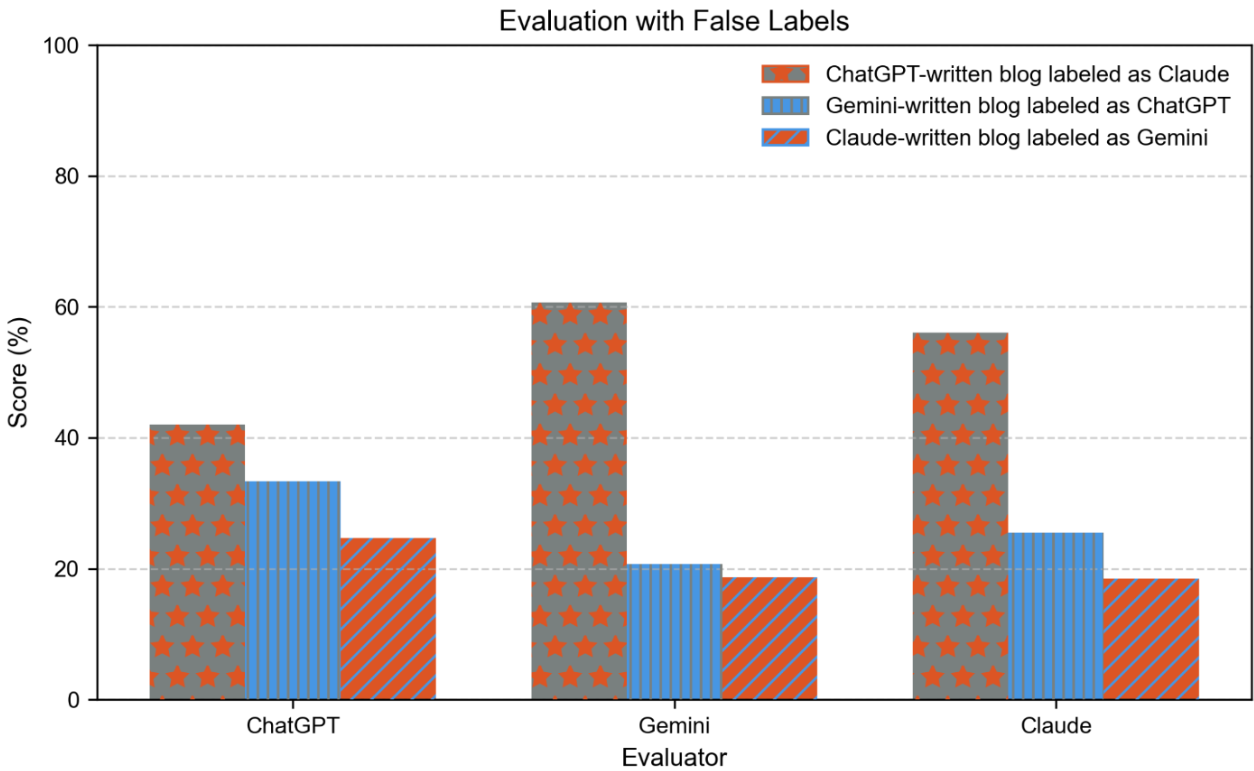}}
\caption{Percentage-based overall scores under \textit{False Label Scenario 2} (ChatGPT-as-Claude, Gemini-as-ChatGPT, Claude-as-Gemini).}
\label{fig}
\end{figure}

\begin{figure}[htbp]
\centerline{\includegraphics[width=1\linewidth]{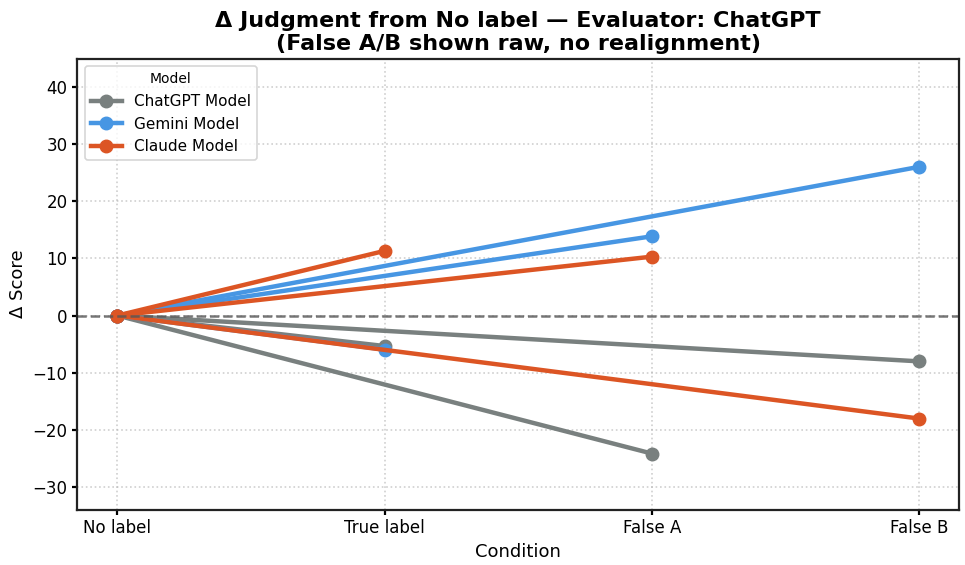}}
\caption{Change of judgment relative to No Label baseline for evaluator \textit{ChatGPT}.}
\label{fig}
\end{figure}
\begin{figure*}[htbp]
\centerline{\includegraphics[width=1\linewidth]{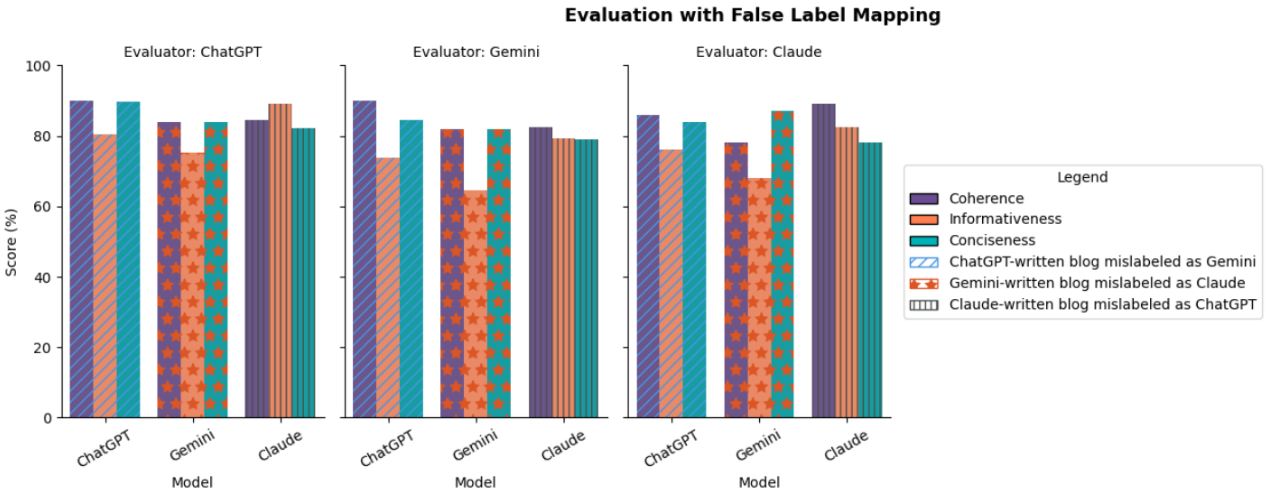}}
\caption{\centering Point-based scores under \textit{False Label Scenario 1}.}
\label{fig}
\end{figure*}

\begin{figure*}[htbp]
\centerline{\includegraphics[width=1\linewidth]{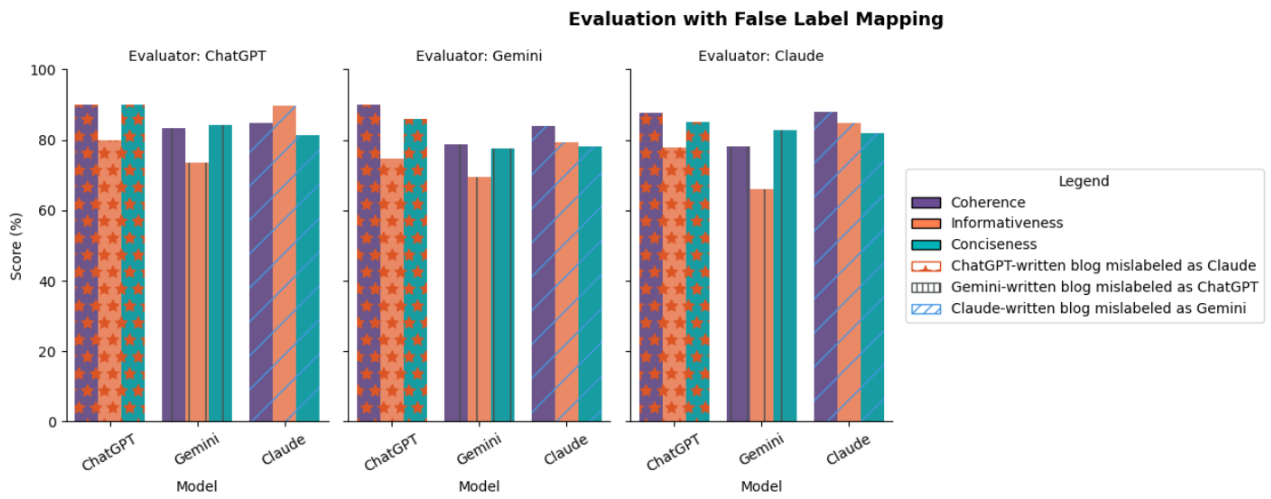}}
\caption{Point-based scores under \textit{False Label Scenario 2}.}
\label{fig}
\end{figure*}

\begin{figure}[htbp]
\centerline{\includegraphics[width=1\linewidth]{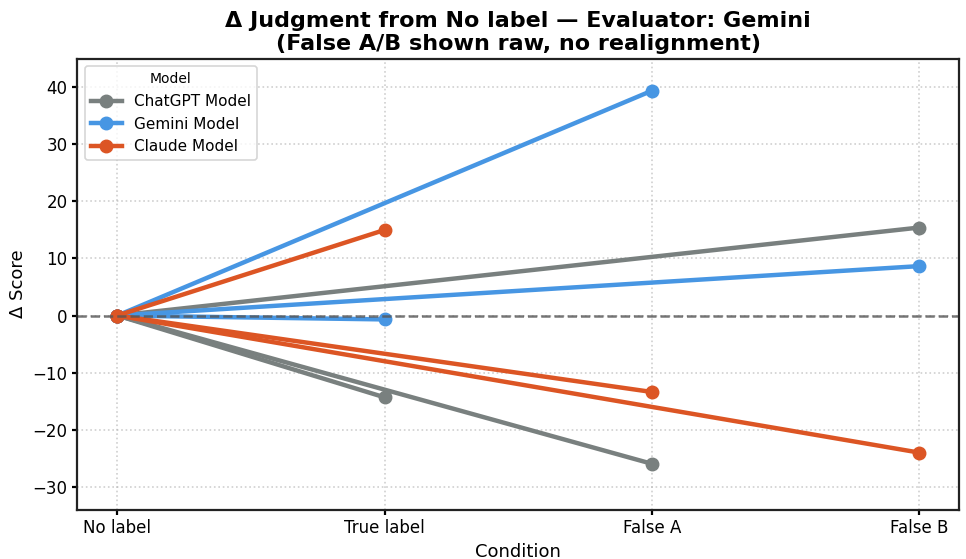}}
\caption{Change of judgment relative to No Label baseline for evaluator \textit{Gemini}.}
\label{fig}
\end{figure}
\begin{figure}[htbp]
\centerline{\includegraphics[width=1\linewidth]{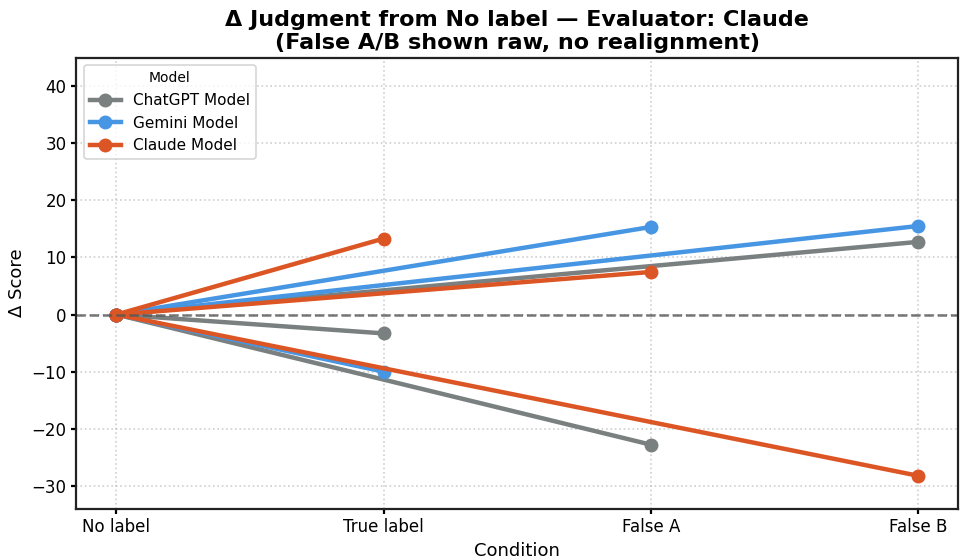}}
\caption{Change of judgment relative to No Label baseline for evaluator \textit{Claude}.}
\label{fig}
\end{figure}

The results are presented in percentage terms for both evaluation formats: (1) overall preference votes and (2) point-based ratings converted to percentages, enabling direct comparison of label effects across evaluation methods.

In the no-label condition, all three models displayed mild self-preference in overall preference scores, with ChatGPT selecting its own outputs 50\% of the time, Gemini 45.3\%, and Claude 46.7\% (Fig. 1). Cross-model scoring revealed consistent undervaluation of Gemini, which received between 7\% and 12\% from other evaluators. Converted quality ratings (Fig. 2) showed the same trend, with Gemini's informativeness averaging 62–78\%, lower than ChatGPT and Claude. Claude consistently achieved the highest coherence ratings (86–89\%) across evaluators, while conciseness remained stable for all models (81–86\%).

Label visibility amplified bias patterns in the true-label condition (Figs. 3–4). Claude's self-score increased to 60\%, receiving the highest preference from all evaluators (54–60\%), while Gemini's scores collapsed to 0\% from Claude, 1.34\% from ChatGPT, and 11.32\% from itself. Converted ratings reflected this Gemini label penalty, with informativeness in the low 60\% range compared to Claude's 79–88\%. ChatGPT's self-preference was moderate at 44.66\%, but it sharply penalized Gemini under true labels.

The false-label conditions revealed the strongest label-induced shifts. In Scenario 1 (ChatGPT-as-Gemini, Gemini-as-Claude, Claude-as-ChatGPT), evaluators consistently favored outputs they believed to be their own, regardless of true authorship (Figs. 5–6). Gemini's preference for content labeled as Claude rose from 11.32\% to 51.35\%, and Claude's preference for content labeled as ChatGPT reached 54.15\%. Informativeness increased by 8–10 percentage points when outputs carried a false "own" label. In Scenario 2 (ChatGPT-as-Claude, Gemini-as-ChatGPT, Claude-as-Gemini), the "Claude" label produced the highest single score, with Gemini awarding 60.7\% to ChatGPT-as-Claude (Figs. 7–8). The "Gemini" label again depressed scores, with Claude-as-Gemini dropping from 60\% under true labels to 18.48\%. "Claude"-labeled outputs achieved top coherence (\~88–89\%), while "Gemini"-labeled outputs received the lowest ratings (\~77–82\%).

Baseline-relative plots (Figs. 9–11) reveal how evaluators' preferences shifted with label introduction. For ChatGPT as evaluator, its own content dropped −5.3 percentage points under true labels while Claude increased +11 points. Under false labels, ChatGPT's content mislabeled as Gemini collapsed −24.2 points, while content labeled as Claude gained +17 points. For Gemini as evaluator, Claude-labeled content received the highest scores (51.35\%–57.68\%), while Gemini-labeled content was severely penalized. For Claude as evaluator, self-preference increased +13.3 points under true labels, while its own content mislabeled as Gemini dropped −28.2 points.

Three consistent trends emerged across conditions. First, the "Claude" label acted as a strong positive bias cue, while the "Gemini" label consistently triggered negative bias. Second, false labels reversed rankings in overall preference scores (swings up to 50 percentage points), whereas point-based ratings showed smaller but aligned shifts (≤12 percentage points). Third, informativeness was the most label-sensitive dimension, while conciseness remained largely stable. Model-specific patterns revealed that Claude showed the strongest self-preference under true labels, Gemini was the most label-sensitive and self-critical, and ChatGPT systematically penalized Gemini outputs under true labels.

\FloatBarrier

\section{Discussion}
The results demonstrate that LLM evaluations are strongly influenced by perceived authorship rather than content alone. The \textit{Claude} label consistently elevated scores regardless of actual authorship, while the \textit{Gemini} label systematically depressed them across all evaluators. This aligns with prior findings on source-based credibility effects in human evaluation [1], suggesting that LLMs may internalize similar biases through training data or alignment tuning.

The false-label scenarios reveal a critical vulnerability: label manipulation can fully reverse evaluation rankings, with preference score shifts reaching 50 percentage points and quality rating shifts up to 12 percentage points. High-level "best choice" judgments proved more susceptible to bias than detailed assessments of coherence, informativeness, and conciseness. Notably, informativeness emerged as the most label-sensitive dimension, while conciseness remained relatively stable, suggesting that structural qualities may be evaluated more objectively than perceived informational value.

Model-specific behaviors highlight distinct evaluation patterns. Claude exhibited the strongest self-preference under true labels (+13 points), while applying severe penalties when its outputs were mislabeled as Gemini (−28 points). Gemini demonstrated harsh self-assessment under true labels but awarded large boosts to "Claude"-labeled text (up to +21 points). ChatGPT showed consistent penalization of Gemini-labeled content across conditions. These asymmetries confirm that model identity can outweigh actual content quality, producing systematic distortions in judgment.

These findings underscore the need for blind evaluation protocols where model identity is hidden, preventing evaluators from anchoring on model names. Multi-model or consensus-based evaluation systems could further mitigate individual biases, particularly for informativeness assessments. Without such safeguards, LLM benchmarking risks overestimating certain models while systematically undervaluing others based on label perception rather than true content quality.

\section{Conclusion}\label{SCM}
This study provides quantitative evidence that LLM evaluations are heavily shaped by label perception, with the "Claude" label consistently boosting and the "Gemini" label consistently depressing scores across evaluators. False labels produced swings of up to 50 percentage points in overall preference, while point-based quality ratings proved more resilient, with changes limited to 12 percentage points and concentrated primarily in informativeness. These findings reveal that perceived model identity can substantially distort judgments independent of content quality.

The results carry critical implications for LLM benchmarking. Blind evaluation protocols, where model identity is concealed, should become standard practice to minimize label-induced bias. Multi-model or consensus-based evaluation systems may further reduce the influence of individual model biases, particularly for subjective dimensions like informativeness. Separating preference judgments from detailed quality ratings could help identify and isolate subjective bias in evaluation frameworks.

Future research should expand this investigation beyond three models to assess whether label effects generalize across a broader set of LLMs and diverse task domains. Investigating the origins of asymmetric label bias whether rooted in training data, alignment procedures, or implicit associations remains essential. Additionally, exploring mitigation strategies such as evaluator fine-tuning, bias-aware scoring adjustments, or adversarial label manipulation could enhance evaluation robustness. By addressing these vulnerabilities, the LLM community can advance toward more reliable, fair, and transparent evaluation standards that prioritize content quality over model reputation.


%



%

%

\end{document}